# Analysis of Kernel Mean Matching under Covariate Shift

Yao-Liang Yu                                                                                   YAOLIANG@CS.UALBERTA.CA
Csaba Szepesvári                                                                               SZEPESVA@CS.UALBERTA.CA
Department of Computing Science, University of Alberta, Edmonton, AB, T6G 2E8, Canada

## Abstract

In real supervised learning scenarios, it is not uncommon that the training and test sample follow different probability distributions, thus rendering the necessity to correct the sampling bias. Focusing on a particular *covariate shift* problem, we derive high probability confidence bounds for the kernel mean matching (KMM) estimator, whose convergence rate turns out to depend on some regularity measure of the regression function and also on some capacity measure of the kernel. By comparing KMM with the natural plug-in estimator, we establish the superiority of the former hence provide concrete evidence/understanding to the effectiveness of KMM under *covariate shift*.

## 1. Introduction

In traditional supervised learning, the training and test sample are usually assumed to be drawn from the *same* probability distribution, however, in practice, this assumption can be easily violated for a variety of reasons, for instance, due to the sampling bias or the nonstationarity of the environment. It is therefore highly desirable to devise algorithms that remain effective under such distribution shifts.

Needless to say the problem is hopeless if the training and test distribution share nothing in common. On the other hand, if the two distributions are indeed related in a nontrivial manner, then it is a quite remarkable fact that effective *adaptation* is possible. Under reasonable assumptions, this problem has been attacked by researchers from statistics (Heckman, 1979; Shimodaira, 2000) and more recently by many researchers from machine learning, see for instance, Zadrozny (2004); Huang et al. (2007); Bickel et al. (2009); Ben-David et al. (2007); Blitzer et al. (2008); Cortes et al. (2008); Sugiyama et al. (2008); Kanamori et al. (2009). We focus in this paper on the *covariate shift* assumption which was first formulated by Shimodaira (2000) and has been followed by many others.

The assumption that the conditional probability distribution of the output variable given the input variable remains fixed in both the training and test set is termed *covariate shift*, *i.e.* the shift happens only for the marginal probability distributions of the covariates. It is well-known that under this setting, the key to correct the sampling bias caused by covariate shift is to estimate the Radon-Nikodym derivative (RND), also called the importance weight or density ratio. A number of methods have been proposed to estimate the RND from finite samples, including kernel mean matching (KMM) (Huang et al., 2007), logistic regression (Bickel et al., 2009), Kullback-Leibler importance estimation (Sugiyama et al., 2008), least-squares (Kanamori et al., 2009), and possibly some others.

Despite of the many algorithms, our current understanding of covariate shift still seems to be limited. From the analyses we are aware of, such as (Gretton et al., 2009) on the confidence bound of the RND by KMM, (Kanamori et al., 2012) on the convergence rate of the least-squares estimate of the RND, and (Cortes et al., 2008) on the distributional stability, they all assume that certain functions lie in the reproducing kernel Hilbert space (RKHS) induced by some user selected kernel. Since this assumption is impossible to verify (even worse, almost certainly violated in practice), one naturally wonders if we can replace it with something more reasonable. Such goal is pursued in this paper and constitutes our main contribution.

We consider the following simple problem: Given the training sample $\{(X_i^{\text{tr}}, Y_i^{\text{tr}})\}_{i=1}^{n_{\text{tr}}}$ and the test sample $\{X_i^{\text{te}}\}_{i=1}^{n_{\text{te}}}$, how well can we estimate the expected value $\mathbb{E}Y^{\text{te}}$, provided *covariate shift* has happened? Note that we do *not* observe the output $Y_i^{\text{te}}$ on the test sample. This problem, at a first glance, ought to be "easy", after all we are humbly asking for estimating a scalar. Indeed, under usual assumptions, plus the nearly impossible assumption that the regression function lies in the RKHS, we prove a parametric rate, that is $\mathcal{O}(n_{\text{tr}}^{-\frac{1}{2}} + n_{\text{te}}^{-\frac{1}{2}})$, for the KMM estimator in Theorem 1 below (to fix ideas, we focus exclusively on KMM in this paper). For a more realistic assumption on the regression function that we borrow from learning theory (Cucker & Zhou, 2007), the convergence rate, proved in Theorem 2, degrades gracefully to $\mathcal{O}(n_{\text{tr}}^{-\frac{\theta}{2(\theta+2)}} + n_{\text{te}}^{-\frac{\theta}{2(\theta+2)}})$, where $\theta > 0$ is a smoothness parameter measuring certain regularity of the regression function (in terms of the kernel). Observe that in the limit when $\theta \to \infty$, the regression function eventually lies in the RKHS and we recover the previous parametric rate. In this regard our bound in Theorem 2 is asymptotically optimal. A very nice feature we discovered for the KMM estimator is that it does not require knowledge of the smoothness parameter $\theta$, thus, it is in some sense *adaptive*.

On the negative side, we show that, if the chosen kernel does not interact very well with the *unknown* regression function, the convergence rate of the KMM estimator could be exceedingly slow, roughly $\mathcal{O}(\log^{-s} \frac{n_{\text{tr}} \cdot n_{\text{te}}}{n_{\text{tr}} + n_{\text{te}}})$, where $s > 0$ again measures certain regularity of the regression function. This unfortunate result should draw attention to the importance of selecting which kernel to be used in practice. A thorough comparison between the KMM estimator and the natural plug-in estimator, conducted in Section 4.3, also reveals the superiority of the former.

We point out that our results are far from giving a complete picture even for the simple problem we consider here, for instance, it is unclear to us whether or not the rate in Theorem 2 can be improved, eventually, to the parametric rate in Theorem 1? Nevertheless, we hope that our paper will convince others about the importance and possibility to work with more reasonable assumptions under *covariate shift*, and as an example, suggest relevant tools which can be used to achieve that goal.

## 2. Preliminaries

In this section we formally state the covariate shift problem under our consideration, followed by some relevant discussions.

### 2.1. Problem Setup

Consider the familiar supervised learning setting, where we are given independent and identically distributed (*i.i.d.*) training samples $\{(X_i^{\text{tr}}, Y_i^{\text{tr}})\}_{i=1}^{n_{\text{tr}}}$ from the joint (Borel) probability measure $\text{P}_{\text{tr}}(\mathrm{d}x, \mathrm{d}y)$ on the (topological) domain $\mathcal{X} \times \mathcal{Y}$, and *i.i.d.* test samples $\{X_i^{\text{te}}\}_{i=1}^{n_{\text{te}}}$ from the joint probability measure $\text{P}_{\text{te}}(\mathrm{d}x, \mathrm{d}y)$ on the same domain. Notice that we do not observe the output $Y_i^{\text{te}}$ on the test sample, and more importantly, we do *not* necessarily assume that the training and test sample are drawn from the *same* probability measure. The problem we consider in this paper is to estimate the expected value $\mathbb{E} Y^{\text{te}}$ from the training sample $\{(X_i^{\text{tr}}, Y_i^{\text{tr}})\}_{i=1}^{n_{\text{tr}}}$ and the test sample $\{X_i^{\text{te}}\}_{i=1}^{n_{\text{te}}}$. In particular, we would like to determine how fast, say, the $1 - \delta$ confidence interval for our estimate shrinks to 0 when the sample sizes $n_{\text{tr}}$ and $n_{\text{te}}$ increase to infinity.

This problem, in its full generality, cannot be solved simply because the training probability measure can be completely irrelevant to the test probability measure that we are interested in. However, if the two probability measures are indeed related in a nontrivial way, our problem becomes solvable. One particular example, which we focus on hereafter, is known in the literature as *covariate shift* (Shimodaira, 2000):

**Assumption 1 (Covariate shift assumption)**

$$\text{P}_{\text{tr}}(\mathrm{d}y|x) = \text{P}_{\text{te}}(\mathrm{d}y|x). \tag{1}$$

We use the same notation for the joint, conditional and marginal probability measures, which should cause no confusion as the arguments would reveal which measure is being referred to. Note that the equality $\text{P}(\mathrm{d}x, \mathrm{d}y) = \text{P}(\mathrm{d}y|x) \cdot \text{P}(\mathrm{d}x)$ holds from the definition of the conditional probability measure, whose existence can be confirmed under very mild assumptions.

Under the covariate shift assumption, the difficulty of our problem, of course, lies entirely on the potential mismatch between the marginal probability measures $\text{P}_{\text{tr}}(\mathrm{d}x)$ and $\text{P}_{\text{te}}(\mathrm{d}x)$. But the Bayes rule already suggests a straightforward approach:

$$\text{P}_{\text{te}}(\mathrm{d}x, \mathrm{d}y) = \text{P}_{\text{te}}(\mathrm{d}y|x) \cdot \text{P}_{\text{te}}(\mathrm{d}x) = \text{P}_{\text{tr}}(\mathrm{d}x, \mathrm{d}y) \cdot \frac{\mathrm{d}\text{P}_{\text{te}}}{\mathrm{d}\text{P}_{\text{tr}}}(x),$$

where the three quantities on the right-hand side can all be estimated from the given samples. However, in order for the above equation to make sense, we need

**Assumption 2 (Continuity assumption)** *The Radon-Nikodym derivative $\beta(x) := \frac{\mathrm{d}\text{P}_{\text{te}}}{\mathrm{d}\text{P}_{\text{tr}}}(x)$ is well-defined and bounded from above by $B < \infty$.*

Note that $B \geq 1$ due to the normalization constraint $\int_{\mathcal{X}} \beta(x) \text{P}_{\text{tr}}(\mathrm{d}x) = 1$. The Radon-Nikodym derivative (RND) is also called the importance weight or the density ratio in the literature. Evidently, if $\beta(x)$ is not well-defined, *i.e.*, there exists some measurable set $A$ such that $P_{te}(A) > 0$ and $P_{tr}(A) = 0$, then in general we cannot infer $P_{te}(dx, dy)$ from merely $P_{tr}(dx), P_{te}(dx)$ and $P_{tr}(dy|x)$, even under the covariate shift assumption. The bounded from above assumption is more artificial. Recently, in a different setting, (Cortes et al., 2010) managed to replace this assumption with a bounded second moment assumption, at the expense of sacrificing the rate a bit. For us, since the domain $\mathcal{X}$ will be assumed to be compact, the bounded from above assumption is not too restrictive (automatically holds when $\beta(x)$ is, say, continuous).

Once we have the RND $\beta(x)$, it becomes easy to correct the sampling bias caused by the mismatch between $P_{tr}(dx)$ and $P_{te}(dx)$, hence solving our problem. Formally, let

$$m(x) := \int_{\mathcal{Y}} y \, P_{te}(dy|x) \qquad (2)$$

be the regression function, then

$$\mathbb{E} Y^{te} = \int_{\mathcal{X}} m(x) \, P_{te}(dx) = \int_{\mathcal{X}} m(x) \beta(x) \, P_{tr}(dx).$$

By the *i.i.d.* assumption, a reasonable estimator for $\mathbb{E} Y^{te}$ would then be $\frac{1}{n_{tr}} \sum_{i=1}^{n_{tr}} \beta(X_i^{tr}) \cdot Y_i^{tr}$. Hence, similarly to most publications on covariate shift, our problem boils down to estimating the RND $\beta(x)$.

### 2.2. A Naive Estimator?

An immediate solution for estimating $\beta(x)$ is to estimate the two marginal measures from the training sample $\{X_i^{tr}\}$ and the test sample $\{X_i^{te}\}$, respectively. For instance, if we know a third (Borel) measure $Q(dx)$ (usually the Lebesgue measure on $\mathbb{R}^d$) such that both $\frac{dP_{te}}{dQ}(x)$ and $\frac{dP_{tr}}{dQ}(x)$ exist, we can employ standard density estimators to estimate them and then set $\hat\beta(x) = \frac{dP_{te}}{dQ}(x) / \frac{dP_{tr}}{dQ}(x)$. However, this naive approach is known to be inferior since density estimation in high dimensions is hard, and moreover, small estimation error in $\frac{dP_{tr}}{dQ}(x)$ could change $\hat\beta(x)$ significantly. To our knowledge, there is little theoretical analysis on this seemingly naive approach.

### 2.3. A Better Estimator?

It seems more appealing to directly estimate the RND $\beta(x)$. Indeed, a large body of work has been devoted to this line of research (Zadrozny, 2004; Huang et al., 2007; Sugiyama et al., 2008; Cortes et al., 2008; Bickel et al., 2009; Kanamori et al., 2009). From the many references, we single out the kernel mean matching (KMM) algorithm, first proposed by Huang et al. (2007) and is also the basis of this paper.

KMM tries to match the mean elements in a feature space induced by a kernel $k(\cdot,\cdot)$ on the domain $\mathcal{X} \times \mathcal{X}$:

$$\min_{\hat\beta_i} \left\{ \hat L(\hat\beta) := \left\| \frac{1}{n_{tr}} \sum_{i=1}^{n_{tr}} \hat\beta_i \Phi(X_i^{tr}) - \frac{1}{n_{te}} \sum_{i=1}^{n_{te}} \Phi(X_i^{te}) \right\|_{\mathcal{H}} \right\}$$
$$\text{s.t.} \quad 0 \leq \hat\beta_i \leq B, \qquad (3)$$

where $\Phi : \mathcal{X} \mapsto \mathcal{H}$ denotes the *canonical* feature map, $\mathcal{H}$ is the reproducing kernel Hilbert space[1] (RKHS) induced by the kernel $k$ and $\|\cdot\|_{\mathcal{H}}$ stands for the norm in $\mathcal{H}$. To simplify later analysis, we have chosen to omit the normalization constraint $\left| \frac{1}{n_{tr}} \sum_{i=1}^{n_{tr}} \hat\beta_i - 1 \right| \leq \epsilon$, where $\epsilon$ is a small positive number, mainly to reflect the fluctuation caused by random samples. It is not hard to verify that (3) is in fact an instance of quadratic programming, hence can be efficiently solved. More details can be found in the paper of Gretton et al. (2009).

A finite sample $1-\delta$ confidence bound for $\hat L(\beta)$ (similar as (10) below) is established in Gretton et al. (2009). This bound is further transferred into a confidence bound for the generalization error of some family of loss minimization algorithms in Cortes et al. (2008), under the notion of distributional stability. However, neither results can provide a direct answer to our problem: a finite sample confidence bound on the estimate of $\mathbb{E} Y^{te}$.

### 2.4. Plug-in Estimator

Another natural approach is to estimate the regression function from the training sample and then plug into the test set. We postpone the discussion and comparison with respect to this estimator until section 4.3.

## 3. Motivation

We motivate the relevance of our problem in this section.

Suppose we have an ensemble of classifiers, say, $\{f_j\}_{j=1}^N$, all trained on the training sample $\{(X_i^{tr}, Y_i^{tr})\}_{i=1}^{n_{tr}}$. A useful task is to compare, hence rank, the classifiers by their generalization errors. This is usually done by assessing the classifiers on some hold out test sample $\{(X_i^{te}, Y_i^{te})\}_{i=1}^{n_{te}}$. It is not uncommon that the test sample is drawn from some different probability measure than the training sample, *i.e.* covariate shift has happened. Since it could be too costly to re-train the classifiers when the test sample is available, we nevertheless still like to have a principled way to rank the classifiers.

---

[1] A thorough background on the theory of reproducing kernels can be found in Aronszajn (1950).

Let $\ell(\cdot,\cdot)$ be the user's favourite loss function, and set $Z_{ij}^{\text{tr}} = \ell(f_j(X_i^{\text{tr}}), Y_i^{\text{tr}})$, $Z_{ij}^{\text{te}} = \ell(f_j(X_i^{\text{te}}), Y_i^{\text{te}})$, then we can use the empirical average of $\{Z_{ij}^{\text{te}}\}_{i=1}^{n_{\text{te}}}$ to estimate the generalization error, that is $\mathbb{E}(Z_{ij}^{\text{te}})$, of classifier $f_j$. But what if we do not have access to $Y_i^{\text{te}}$ hence consequently $Z_{ij}^{\text{te}}$? Can we still accomplish the ranking job?

The answer is yes, and it is precisely the covariate shift problem under our consideration. To see that, consider the pair $\{X_i^{\text{tr}}, Z_{ij}^{\text{tr}}\}_{i=1}^{n_{\text{tr}}}$ and $\{X_i^{\text{te}}\}_{i=1}^{n_{\text{te}}}$. Under the covariate shift assumption, that is $P_{\text{tr}}(dy|x) = P_{\text{te}}(dy|x)$, it is not hard to see that $P_{\text{tr}}(dz|x) = P_{\text{te}}(dz|x)$, hence the covariate shift assumption holds for the ranking problem, therefore the confidence bounds derived in the next section provide an effective solution.

We do not report numerical experiments in this paper for two reasons: 1). Our main interest is on theoretical analysis; 2). Exhaustive experimental results on KMM can already be found in Gretton et al. (2009).

## 4. Theoretical Analysis

This section contains our main contribution, *i.e.*, a theoretical analysis of the KMM estimator for $\mathbb{E}Y^{\text{te}}$.

### 4.1. The population version

Let us first take a look at the population version of KMM[2], which is much easier to analyze and provides valuable insights:

$$\hat{\beta}^* \in \arg\min_{\hat{\beta}} \left\| \int_{\mathcal{X}} \Phi(x)\hat{\beta}(x)P_{\text{tr}}(dx) - \int_{\mathcal{X}} \Phi(x)P_{\text{te}}(dx) \right\|_{\mathcal{H}}$$
$$\text{s.t.} \quad 0 \leq \hat{\beta} \leq B, \quad \int_{\mathcal{X}} \hat{\beta}(x)P_{\text{tr}}(dx) = 1.$$

The minimum value is 0 since the true RND $\beta(x)$ is apparently feasible, hence at optimum we always have

$$\int_{\mathcal{X}} \Phi(x)\hat{\beta}^*(x)P_{\text{tr}}(dx) = \int_{\mathcal{X}} \Phi(x)P_{\text{te}}(dx). \quad (4)$$

The question is whether the natural estimator $\int_{\mathcal{X} \times \mathcal{Y}} \hat{\beta}^*(x)y\, P_{\text{tr}}(dx, dy)$ is consistent? In other words, is

$$\int_{\mathcal{X}} m(x)\hat{\beta}^*(x)P_{\text{tr}}(dx) \stackrel{?}{=} \mathbb{E}Y^{\text{te}} = \int_{\mathcal{X}} m(x)\beta(x)P_{\text{tr}}(dx), \quad (5)$$

---
[2]All Hilbert space valued integrals in this paper are to be understood as the Bochner integral (Yosida, 1980).

where recall that $m(x)$ is the regression function defined in (2) and $\beta(x)$ is the true RND.

The equality in (5) indeed holds under at least two conditions (respectively). First, if the regression function $m \in \mathcal{H}$, then taking inner products with $m$ in (4) and applying the reproducing property we get (5). Second, if the kernel $k$ is *characteristic* (Sriperumbudur et al., 2010), meaning that the map $\int_{\mathcal{X}} \Phi(x)P(dx)$ from the space of probability measures to the RKHS $\mathcal{H}$ is injective, then we conclude $\hat{\beta}^* = \beta$ from (4) hence follows (5).

The above two cases suggest the possibility of solving our problem by KMM. Of course, in reality one only has finite samples from the underlying probability measures, thus calls for a thorough study of the empirical KMM, *i.e.* (3). Interestingly, our analysis reveals that in the first case above, we indeed can have a parametric rate while in the second case the rate becomes nonparametric, hence inferior (but does not seem to rely on the *characteristic* property of the kernel).

### 4.2. The empirical version

In this subsection we analyze KMM in details. The following assumption will be needed:

**Assumption 3 (Compactness assumption)** $\mathcal{X}$ *is a compact metrizable space, $\mathcal{Y} \subseteq [0,1]$, and the kernel $k$ is continuous, whence $\|k\|_\infty \leq C^2 < \infty$.*

We use $\|\cdot\|_\infty$ for the supremum norm. Under the above assumption, the feature map $\Phi$ is continuous hence measurable (with respect to the Borel $\sigma$-fields), and the RKHS is separable, therefore the Bochner integrals in the previous subsection are well-defined. Moreover, the conditional probability measure indeed exists under our assumption.

We are now ready to derive a finite sample confidence bound for our estimate $|\frac{1}{n_{\text{tr}}} \sum_{i=1}^{n_{\text{tr}}} \hat{\beta}_i Y_i^{\text{tr}} - \mathbb{E}Y^{\text{te}}|$, where $\hat{\beta}_i$ is a minimizer of (3). We start by splitting the sum:

$$\frac{1}{n_{\text{tr}}} \sum_{i=1}^{n_{\text{tr}}} \hat{\beta}_i Y_i^{\text{tr}} - \mathbb{E}Y^{\text{te}} = \frac{1}{n_{\text{tr}}} \sum_{i=1}^{n_{\text{tr}}} \hat{\beta}_i(Y_i^{\text{tr}} - m(X_i^{\text{tr}}))$$
$$+ \frac{1}{n_{\text{tr}}} \sum_{i=1}^{n_{\text{tr}}} (\hat{\beta}_i - \beta_i)(m(X_i^{\text{tr}}) - h(X_i^{\text{tr}}))$$
$$+ \frac{1}{n_{\text{tr}}} \sum_{i=1}^{n_{\text{tr}}} (\hat{\beta}_i - \beta_i)h(X_i^{\text{tr}})$$
$$+ \frac{1}{n_{\text{tr}}} \sum_{i=1}^{n_{\text{tr}}} \beta_i m(X_i^{\text{tr}}) - \mathbb{E}Y^{\text{te}}, \quad (6)$$

where $\beta_i := \beta(X_i^{\text{tr}})$ and $h \in \mathcal{H}$ is to be specified later.

Analysis of Kernel Mean Matching under Covariate ShiftWe bound each term individually. For the last term in (6), we can apply Hoeffding's inequality (Hoeffding, 1963) to conclude that with probability at least $1 - \delta$,

$$\left|\frac{1}{n_{\text{tr}}} \sum_{i=1}^{n_{\text{tr}}} \beta_i m(X_i^{\text{tr}}) - \mathbb{E} Y^{\text{te}}\right| \leq B \sqrt{\frac{1}{2n_{\text{tr}}} \log \frac{2}{\delta}}. \quad (7)$$

The first term in (6) can be bounded similarly. Conditioned on $\{X_i^{\text{tr}}\}$ and $\{X_i^{\text{te}}\}$, we apply again Hoeffding's inequality. Note that $\hat{\beta}_i(Y_i^{\text{tr}} - m(X_i^{\text{tr}})) \in [-\hat{\beta}_i m(X_i^{\text{tr}}), \hat{\beta}_i(1 - m(X_i^{\text{tr}}))]$, therefore its range is of size $\hat{\beta}_i$. With probability at least $1 - \delta$,

$$\left|\frac{1}{n_{\text{tr}}} \sum_{i=1}^{n_{\text{tr}}} \hat{\beta}_i(Y_i^{\text{tr}} - m(X_i^{\text{tr}}))\right| \leq \sqrt{\frac{1}{n_{\text{tr}}} \sum_{i=1}^{n_{\text{tr}}} \hat{\beta}_i^2} \cdot \sqrt{\frac{1}{2n_{\text{tr}}} \log \frac{2}{\delta}}$$
$$\leq B \sqrt{\frac{1}{2n_{\text{tr}}} \log \frac{2}{\delta}}. \quad (8)$$

The second and third terms in (6) require more work. Consider first the third term:

$$\left|\frac{1}{n_{\text{tr}}} \sum_{i=1}^{n_{\text{tr}}} (\hat{\beta}_i - \beta_i) h(X_i^{\text{tr}})\right| = \left|\frac{1}{n_{\text{tr}}} \sum_{i=1}^{n_{\text{tr}}} (\hat{\beta}_i - \beta_i) \langle h, \Phi(X_i^{\text{tr}}) \rangle\right|$$
$$\leq \|h\|_{\mathcal{H}} \cdot \left\|\frac{1}{n_{\text{tr}}} \sum_{i=1}^{n_{\text{tr}}} (\hat{\beta}_i - \beta_i) \Phi(X_i^{\text{tr}})\right\|_{\mathcal{H}}$$
$$\leq \|h\|_{\mathcal{H}} \cdot [\hat{L}(\hat{\beta}) + \hat{L}(\beta_{1:n_{\text{tr}}})]$$
$$\leq \|h\|_{\mathcal{H}} \cdot 2\hat{L}(\beta_{1:n_{\text{tr}}}), \quad (9)$$

where $\beta_{1:n_{\text{tr}}}$ denotes the restriction of $\beta$ to the training sample $\{X_i^{\text{tr}}\}$, $\hat{L}(\cdot)$ is defined in (3), and the equality is because $h \in \mathcal{H}$ (and the reproducing property of the *canonical* feature map), the first inequality is by the Cauchy-Schwarz inequality, the second inequality is due to the triangle inequality, and the last inequality is by the optimality of $\hat{\beta}$ and the feasibility of $\beta_{1:n_{\text{tr}}}$ in problem (3). Next, we bound $\hat{L}(\beta_{1:n_{\text{tr}}})$:

$$\hat{L}(\beta_{1:n_{\text{tr}}}) := \left\|\frac{1}{n_{\text{tr}}} \sum_{i=1}^{n_{\text{tr}}} \beta_i \Phi(X_i^{\text{tr}}) - \frac{1}{n_{\text{te}}} \sum_{i=1}^{n_{\text{te}}} \Phi(X_i^{\text{te}})\right\|_{\mathcal{H}}$$
$$\leq C \sqrt{2\left(\frac{B^2}{n_{\text{tr}}} + \frac{1}{n_{\text{te}}}\right) \log \frac{2}{\delta}} \quad (10)$$

with probability at least $1 - \delta$, where the inequality follows from the Hilbert space valued Hoeffding inequality in (Pinelis, 1994, Theorem 3.5). Note that Pinelis proved his inequality for martingales in any 2-smooth separable Banach space (Hilbert spaces are *bona fide* 2-smooth). We remark that another way, see for instance (Gretton et al., 2009, Lemma 1.5), is to use McDiarmid's inequality to bound $\hat{L}(\beta_{1:n_{\text{tr}}})$ by its expectation, and then bound the expectation straightforwardly. In general, Pinelis's inequality will lead to (slightly) tighter bounds due to its known optimality (in certain sense).

Finally, we come to the second term left in (6), which is roughly the approximation error in learning theory (Cucker & Zhou, 2007). Note that all confidence bounds we have derived so far shrink at the parametric rate $\mathcal{O}(\sqrt{1/n_{\text{tr}} + 1/n_{\text{te}}})$. However, from here on we will have to tolerate nonparametric rates. Since we are going to apply different approximation error bounds to the second term in (6), it seems more convenient to collect the results separately. We start with an encouraging result:

**Theorem 1** *Under Assumptions 1-3, if the regression function $m \in \mathcal{H}$ (the RKHS induced by the kernel $k$), then with probability at least[3] $1 - \delta$,*

$$\left|\frac{1}{n_{\text{tr}}} \sum_{i=1}^{n_{\text{tr}}} \hat{\beta}_i Y_i^{\text{tr}} - \mathbb{E} Y^{\text{te}}\right| \leq M \cdot \sqrt{2\left(\frac{B^2}{n_{\text{tr}}} + \frac{1}{n_{\text{te}}}\right) \log \frac{6}{\delta}},$$

*where $M := 1 + 2C\|m\|_{\mathcal{H}}$ and $\hat{\beta}_i$ is computed from (3).*

*Proof:* By assumption, setting $h = m$ zeros out the second term in (6). A standard union bound combining (7)-(10) completes the proof (and we simplified the bound by slightly worsening the constant). ∎

The confidence bound shrinks at the parametric rate, although the constant depends on $\|m\|_{\mathcal{H}}$, which in general is not computable, but can be estimated from the training sample $\{(X_i^{\text{tr}}, Y_i^{\text{tr}})\}$ at a rate worse than parametric. Since this estimate inevitably introduces other uncomputable quantities, we omit the relevant discussion. On the other hand, our bound suggests that if *a priori* information about $m$ is indeed available, one should choose a kernel that minimizes its induced norm on $m$.

The case when $m \notin \mathcal{H}$ is less satisfactory, despite of its practicality. We point out that a denseness argument cannot resolve this difficulty. To be more precise, let us assume for a moment $m \in \mathscr{C}(\mathcal{X})$ (the space of continuous functions on $\mathcal{X}$) and $k$ be a *universal* kernel (Steinwart, 2002), meaning that the RKHS induced by $k$ is dense in $(\mathscr{C}(\mathcal{X}), \|\cdot\|_\infty)$. By the assumed universal property of the kernel, there exists suitable $h \in \mathcal{H}$ that makes the second term in (6) arbitrarily small (in fact, can be made vanishing), however, on the other hand, recall that the bound (9) on the third term in (6) depends on $\|h\|_{\mathcal{H}}$ hence could blow up. If we trade off the two terms appropriately, we might get a rate that is acceptable (but worse than parametric). The next theorem concretizes this idea.

---

[3] Throughout this paper, the confidence parameter $\delta$ is always taken *arbitrarily* in $(0, 1)$.

**Theorem 2** *Under Assumptions 1-3, if $\mathcal{A}_2(m, R) := \inf_{\|g\|_\mathcal{H} \leq R} \|m - g\|_{\mathscr{L}^2_{P_{tr}}} \leq C_2 R^{-\theta/2}$ for some $\theta > 0$ and constant $C_2 \geq 0$, then with probability at least $1 - \delta$,*

$$\left|\frac{1}{n_{tr}}\sum_{i=1}^{n_{tr}}\hat{\beta}_i Y_i^{tr} - \mathbb{E}Y^{te}\right| \leq$$
$$B\sqrt{\frac{9}{2n_{tr}}\log\frac{8}{\delta}} + C_\theta(BC_2)^{\frac{2}{\theta+2}} D_2^{\frac{\theta}{\theta+2}},$$

*where $D_2 := 2C\sqrt{2\left(\frac{B^2}{n_{tr}} + \frac{1}{n_{te}}\right)\log\frac{8}{\delta}} + BC\sqrt{\frac{1}{2n_{tr}}\log\frac{8}{\delta}}$, $C_\theta := (1 + 2/\theta)\left(\frac{\theta}{2}\right)^{\frac{2}{\theta+2}}$ and $\hat{\beta}_i$ is computed from (3).*

*Proof:* By the triangle inequality,

$$\left|\frac{1}{n_{tr}}\sum_{i=1}^{n_{tr}}(\hat{\beta}_i - \beta_i)(m(X_i^{tr}) - h(X_i^{tr}))\right|$$
$$\leq B \cdot \frac{1}{n_{tr}}\sum_{i=1}^{n_{tr}}|m(X_i^{tr}) - h(X_i^{tr})|.$$

Not surprisingly, we apply yet again Hoeffding's inequality to relate the last term above to its expectation. Since

$$\|m - h\|_\infty \leq 1 + \|\langle h, \Phi(\cdot)\rangle\|_\infty \leq 1 + C\|h\|_\mathcal{H},$$

we have with probability at least $1 - \delta$,

$$\frac{1}{n_{tr}}\sum_{i=1}^{n_{tr}}|m(X_i^{tr}) - h(X_i^{tr})| \leq (1 + CR)\sqrt{\frac{1}{2n_{tr}}\log\frac{2}{\delta}} + \mathcal{A}_2(m, R),$$

where $R := \|h\|_\mathcal{H}$. Combining this bound with (7)-(10) and applying our assumption on $\mathcal{A}_2(m, R)$:

$$\left|\frac{1}{n_{tr}}\sum_{i=1}^{n_{tr}}(\hat{\beta}_i - \beta_i)(m(X_i^{tr}) - h(X_i^{tr}))\right|$$
$$\leq B\sqrt{\frac{2}{n_{tr}}\log\frac{8}{\delta}} + 2RC\sqrt{2\left(\frac{B^2}{n_{tr}} + \frac{1}{n_{te}}\right)\log\frac{8}{\delta}}$$
$$+ BC_2 R^{-\theta/2} + B(1 + CR)\sqrt{\frac{1}{2n_{tr}}\log\frac{8}{\delta}}.$$

Setting $R = \left(\frac{\theta BC_2}{2D_2}\right)^{\frac{2}{\theta+2}}$ completes the proof. ∎

In Theorem 2 we do not even assume $m \in \mathscr{C}(\mathcal{X})$; all we need is $m \in \mathscr{L}^2_{P_{tr}}$, the space of $P_{tr}(dx)$ square integrable functions. The latter condition always holds since $0 \leq m \leq 1$ by Assumption 3. The quantity $\mathcal{A}_2(m, R)$ is called the approximation error in learning theory and its polynomial decay is known to be (almost) equivalent to $m \in \text{Range}(\mathcal{T}_k^{\frac{\theta}{2\theta+4}})$, see for instance Theorem 4.1 of Cucker & Zhou (2007). Here $\mathcal{T}_k$ is the integral operator $(\mathcal{T}_k f)(x') = \int_\mathcal{X} k(x', x)f(x)P_{tr}(dx)$ on $\mathscr{L}^2_{P_{tr}}$. The smoothness parameter $\theta > 0$ measures the *regularity* of the regression function, and as it increases, the range space of $\mathcal{T}_k^{\frac{\theta}{2\theta+4}}$ becomes smaller, hence our decay assumption on $\mathcal{A}_2(m, R)$ becomes more stringent. Note that the exponent $\frac{\theta}{2\theta+4}$ is necessarily smaller than $1/2$ (but approaches $1/2$ when $\theta \to \infty$) because by Mercer's theorem $\mathcal{T}_k^{\frac{1}{2}}$ is onto $\mathcal{H}$ (in which case the range assumption would bring us back to Theorem 1).

Theorem 2 shows that the confidence bound now shrinks at a slower rate, roughly $\mathcal{O}(n_{tr}^{-\frac{\theta}{2(\theta+2)}} + n_{te}^{-\frac{\theta}{2(\theta+2)}})$, which, as $\theta \to \infty$, approaches the parametric rate $\mathcal{O}(n_{tr}^{-\frac{1}{2}} + n_{te}^{-\frac{1}{2}})$ derived in Theorem 1 where we assume $m \in \mathcal{H}$. We point out that the source of this slower rate comes from the irregular nature of the regression function (in the eye of the kernel $k$).

The polynomial decay assumption on $\mathcal{A}_2(m, R)$ is not always satisfied, for instance, it is shown in Theorem 6.2 of Cucker & Zhou (2007) that for $\mathscr{C}^\infty$ (indefinite times differentiable) kernels (such as the popular Gaussian kernel), polynomial decay implies that the regression function $m \in \mathscr{C}^\infty(\mathcal{X})$ (under mild assumptions on $\mathcal{X}$ and $P_{tr}(dx)$). Therefore, as long as one works with smooth kernels but nonsmooth regression functions, the approximation error has to decay logarithmically slowly. We give a logarithmic bound for such cases.

**Theorem 3** *Under Assumptions 1-3, if $\mathcal{A}_\infty(m, R) := \inf_{\|g\|_\mathcal{H} \leq R} \|m - g\|_\infty \leq C_\infty(\log R)^{-s}$ for some $s > 0$ and constant $C_\infty \geq 0$ (assuming $R \geq 1$), then (for $n_{tr}$ and $n_{te}$ larger than some constant),*

$$\left|\frac{1}{n_{tr}}\sum_{i=1}^{n_{tr}}\hat{\beta}_i Y_i^{tr} - \mathbb{E}Y^{te}\right| \leq \left(1 + \frac{1}{s}\right)^s BC_\infty \left(\log\frac{sBC_\infty}{D_\infty}\right)^{-s}$$
$$+ B\sqrt{\frac{2}{n_{tr}}\log\frac{6}{\delta}} + (sBC_\infty)^{\frac{s}{s+1}} D_\infty^{\frac{1}{s+1}}$$

*holds with probability at least $1 - \delta$, where $D_\infty = 2C\sqrt{2\left(\frac{B^2}{n_{tr}} + \frac{1}{n_{te}}\right)\log\frac{6}{\delta}}$ and $\hat{\beta}_i$ is computed from (3).*

The proof is similar as that of Theorem 2 except that we set $R = \left(\frac{sBC_\infty}{D_\infty}\right)^{\frac{s}{s+1}}$.

Theorem 3 shows that in such unfavourable cases, the confidence bound shrinks at an exceedingly slow rate, roughly, $\mathcal{O}(\log^{-s}\frac{n_{\text{tr}}\cdot n_{\text{te}}}{n_{\text{tr}}+n_{\text{te}}})$. The reason, of course, is due to the slow decay of the approximation error $\mathcal{A}_\infty(m,R)$. It is proved in Theorem 6.1 of Cucker & Zhou (2007) that for the Gaussian kernel $k(x',x) = \exp(-\|x-x'\|_2^2/\sigma^2)$, if $\mathcal{X} \subseteq \mathbb{R}^d$ has smooth boundary and the regression function $m \in H^s(\mathcal{X})$ with index $s > d/2$, then the logarithmic decay assumed in Theorem 3 holds. Here $H^s(\mathcal{X})$ is the Sobolev space (the completion of $\mathscr{C}^\infty(\mathcal{X})$ under the inner product $\langle f,g \rangle_s := \int_\mathcal{X} \sum_{|\alpha|\le s} \frac{\mathrm{d}^\alpha f}{\mathrm{d}x}\frac{\mathrm{d}^\alpha g}{\mathrm{d}x}$, assuming $s \in \mathbb{N}$). Similar bounds also hold for the inverse multiquadrics kernel $k(x',x) = (c^2 + \|x-x'\|_2^2)^{-\alpha}$ with $\alpha > 0$. We remark that in this regard Theorem 3 disrespects the popular Gaussian kernel used ubiquitously in practice and should draw the attention of researchers.

### 4.3. Discussion

It seems worthwhile to devote a subsection to discussing a very natural question that the reader might already have: why not estimate the regression function $m$ on the training set and then plug into the test set, after all $m$ does not change under the covariate shift assumption? Algorithmically, this is perfectly doable, perhaps conceptually even simpler since the algorithm does not need to see the test data beforehand. We note that estimating the regression function from *i.i.d.* samples has been well studied in the learning theory literature, see for instance, Chapter 8 of Cucker & Zhou (2007) and the many references therein.

The difficulty, though, lies in the appropriate error metric on the estimate. Recall that when estimating the regression function from *i.i.d. training* samples, one usually measures the progress (*i.e.* the discrepancy between the estimate $\hat{m}$ and $m$) by the $\mathscr{L}^2$ norm under the *training* probability measure $\mathrm{P}_{\text{tr}}(\mathrm{d}x)$, while what we really want is a confidence bound on the term

$$\left| \frac{1}{n_{\text{te}}} \sum_{i=1}^{n_{\text{te}}} \hat{m}(X_i^{\text{te}}) - \mathbb{E}Y^{\text{te}} \right|. \quad (11)$$

Since $\mathrm{P}_{\text{tr}} \ne \mathrm{P}_{\text{te}}$, there is evidently a probability measure mismatch between the bound we have from estimating $m$ and the true interested quantity. Indeed, conditioned on the training sample $\{(X_i^{\text{tr}}, Y_i^{\text{tr}})\}$, using the triangle inequality we can bound (11) by :

$$\left| \frac{1}{n_{\text{te}}} \sum_{i=1}^{n_{\text{te}}} \hat{m}(X_i^{\text{te}}) - \int \hat{m}(x)\mathrm{P}_{\text{te}}(\mathrm{d}x) \right| + \|\hat{m}-m\|_{\mathscr{L}^2_{\mathrm{P}_{\text{te}}}}.$$

The first term above can be bounded again through Hoeffding's inequality, while the second term is close to what we usually have from estimating $m$: the only difference being that the $\mathscr{L}^2$ norm is now under the *test* probability measure $\mathrm{P}_{\text{te}}(\mathrm{d}x)$. Fortunately, since the norm of the identity map $\text{id} : ([-1,1]^\mathcal{X}, \|\cdot\|_{\mathscr{L}^2_{\mathrm{P}_{\text{tr}}}}) \mapsto ([-1,1]^\mathcal{X}, \|\cdot\|_{\mathscr{L}^2_{\mathrm{P}_{\text{te}}}})$ is bounded by $\sqrt{B}$ (see Assumption 2), we can deduce a bound for (11) based upon results from estimating $m$, though less appealingly, a much looser bound than the one given in Theorem 2. We record such a result for the purpose of comparison:

**Theorem 4** *Under Assumptions 1-3, if the regression function* $m \in \text{Range}(\mathcal{T}_k^{\frac{\theta}{2\theta+4}})$ *for some* $\theta > 0$, *then with probability at least* $1-\delta$,

$$\left| \frac{1}{n_{\text{te}}} \sum_{i=1}^{n_{\text{te}}} \hat{m}(Y_i^{\text{te}}) - \mathbb{E}Y^{\text{te}} \right| \le \sqrt{\frac{1}{2n_{\text{te}}}\log\frac{4}{\delta}} + \sqrt{B}C_1 n_{\text{tr}}^{-\frac{3\theta}{12\theta+16}},$$

*where $C_1$ is some constant that does not depend on $n_{\text{tr}}, n_{\text{te}}$, and $\hat{m}$ is the (regularized least-squares) estimate of $m$ in Smale & Zhou (2007).*

The theorem follows from the bound on $\|\hat{m}-m\|_{\mathscr{L}^2_{\mathrm{P}_{\text{tr}}}}$ in Corollary 3.2 of Sun & Wu (2009), which is an improvement over Smale & Zhou (2007).

Carefully comparing the current theorem with Theorem 2, we observe: 1). Theorem 4, which is based on the regularized least-squares estimate of the regression function, needs to know in advance the parameter $\theta$ (in order to tune the regularization constant) while Theorem 2, derived for KMM, does *not* require any such information, hence in some sense KMM is "adaptive"; 2). Theorem 4 has much worse dependence on the training sample size $n_{\text{tr}}$; it does not recover the parametric rate even when the smoothness parameter $\theta$ goes to $\infty$ (we get $n_{\text{tr}}^{-1/4}$, instead of $n_{\text{tr}}^{-1/2}$). On the other hand, Theorem 4 has better dependence on the test sample size $n_{\text{te}}$, which is, however, probably not so important since usually one has much more test samples than training samples because the lack of labels make the former much easier to acquire; 3). Theorem 4 seems to have better dependence on the parameter $B$; 4). Given the fact that KMM utilizes both the training data and the test data in the learning phase, it is not entirely a surprise that KMM wins in terms of convergence rate, nevertheless, we find it quite stunning that by sacrificing the rate slightly on $n_{\text{te}}$, KMM is able to improve the rate on $n_{\text{tr}}$ so significantly.

## 5. Conclusion

For estimating the expected value of the output on the test set where *covariate shift* has happened, we have derived high probability confidence bounds for the kernel mean matching (KMM) estimator, which converges, roughly $\mathcal{O}(n_{\text{tr}}^{-\frac{1}{2}} + n_{\text{te}}^{-\frac{1}{2}})$ when the regression function lies in the RKHS, and more generally $\mathcal{O}(n_{\text{tr}}^{-\frac{\theta}{2(\theta+2)}} + n_{\text{te}}^{-\frac{\theta}{2(\theta+2)}})$ when the regression function exhibits certain regularity measured by $\theta$. An extremely slow rate, roughly $\mathcal{O}(\log^{-s} \frac{n_{\text{tr}} \cdot n_{\text{te}}}{n_{\text{tr}} + n_{\text{te}}})$, is also provided, calling attention of choosing the right kernel. From the comparison of the bounds, KMM proves to be much more superior than the plug-in estimator hence provides concrete evidence/understanding to the effectiveness of KMM under *covariate shift*.

Although it is unclear to us if it is possible to avoid approximating the regression function, we suspect the bound in Theorem 2 is in some sense optimal and we are currently investigating it. We also plan to generalize our results to the least-squares estimation problem.

## Acknowledgements

This work was supported by Alberta Innovates Technology Futures and NSERC.

## References

Aronszajn, Nachman. Theory of reproducing kernels. *Transactions of the American Mathematical Sociery*, 68:337–404, 1950.

Ben-David, Shai, Blitzer, John, Crammer, Koby, and Pereira, Fernando. Analysis of representations for domain adaptation. In *NIPS*, pp. 137–144. 2007.

Bickel, Steffen, Brückner, Michael, and Scheffer, Tobias. Discriminative learning under covariate shift. *JMLR*, 10:2137–2155, 2009.

Blitzer, John, Crammer, Koby, Kulesza, Alex, Pereira, Fernando, and Wortman, Jennifer. Learning bounds for domain adaptation. In *NIPS*, pp. 129–136. 2008.

Cortes, Corinna, Mohri, Mehryar, Riley, Michael, and Rostamizadeh, Afshin. Sample selection bias correction theory. In *ALT*, pp. 38–53. 2008.

Cortes, Corinna, Mansour, Yishay, and Mohri, Mehryar. Learning bounds for importance weighting. In *NIPS*, pp. 442–450. 2010.

Cucker, Felipe and Zhou, Ding-Xuan. *Learning theory: an approximation theory viewpoint*. Cambridge University Press, 2007.

Gretton, Arthur, Smola, Alexander J., Huang, Jiayuan, Schmittfull, Marcel, Borgwardt, Karsten M., and Schölkopf, Bernhard. *Covariate Shift by Kernel Mean Matching*, pp. 131–160. MIT Press, 2009.

Heckman, James J. Sample selection bias as a specification error. *Econometrica*, 47(1):153–161, 1979.

Hoeffding, Wassily. Probability inequalities for sums of bounded random variables. *Journal of the American Statistical Association*, 58(301):13–30, 1963.

Huang, Jiayuan, Smola, Alexander J., Gretton, Arthur, Borgwardt, Karsten M., and Schölkopf, Bernhard. Correcting sample selection bias by unlabeled data. In *NIPS*, pp. 601–608. 2007.

Kanamori, Takafumi, Hido, Shohei, and Sugiyama, Masashi. A least-squares approach to direct importance estimation. *JMLR*, 10:1391–1445, 2009.

Kanamori, Takafumi, Suzuki, Taiji, and Sugiyama, Masashi. Statistical analysis of kernel-based least-squares density-ratio estimation. *Machine Learning*, 86:335–367, 2012.

Pinelis, Iosif. Optimum bounds for the distributions of martingales in Banach spaces. *The Annals of Probability*, 22(4):1679–1706, 1994.

Shimodaira, Hidetoshi. Improving predictive inference under covariate shift by weighting the log-likelihood function. *Journal of Statistical Planning and Inference*, 90(2):227–244, 2000.

Smale, Steve and Zhou, Ding-Xuan. Learning theory estimates via integral operators and their approximations. *Constructive Approximation*, 26:153–172, 2007.

Sriperumbudur, Bharath K., Gretton, Arthur, Fukumizu, Kenji, Schölkopf, Bernhard, and Lanckriet, Gert R. G. Hilbert space embeddings and metrics on probability measures. *JMLR*, 11:1517–1561, 2010.

Steinwart, Ingo. On the influence of the kernel on the consistency of support vector machines. *JMLR*, 2:67–93, 2002.

Sugiyama, Masashi, Nakajima, Shinichi, Kashima, Hisashi, Buenau, Paul Von, and Kawanabe, Motoaki. Direct importance estimation with model selection and its application to covariate shift adaptation. In *NIPS*, pp. 1433–1440. 2008.

Sun, Hongwei and Wu, Qiang. A note on application of integral operator in learning theory. *Applied and Computational Harmonic Analysis*, 26:416–421, 2009.

Yosida, Kôsaku. *Functional Analysis*. Springer, 6th edition, 1980.

Zadrozny, Bianca. Learning and evaluating classifiers under sample selection bias. In *ICML*. 2004.